%% file: main.tex
\documentclass[runningheads]{llncs}
\usepackage[T1]{fontenc}
\usepackage{graphicx}
\usepackage{multirow}
\usepackage{booktabs}
\usepackage{amsfonts}
\usepackage{ctable}
\usepackage{color, colortbl, xcolor, soul}
\usepackage[colorlinks=true,linkcolor=blue, citecolor=blue, urlcolor=blue]{hyperref}
\usepackage{pifont}
\usepackage[ruled, vlined]{algorithm2e}
\let\oldnl\nl
\newcommand\scalemath[2]{\scalebox{#1}{\mbox{\ensuremath{\displaystyle #2}}}}
\newcommand{\nonl}{\renewcommand{\nl}{\let\nl\oldnl}}
\usepackage{etoolbox}
\usepackage{xr}
\makeatletter
\patchcmd\algocf@Vline{\vrule}{\vrule \kern-0.4pt}{}{}
\patchcmd\algocf@Vsline{\vrule}{\vrule \kern-0.4pt}{}{}
\makeatother

\definecolor{Gray1}{gray}{0.94}
\definecolor{Gray2}{gray}{0.84}
\sethlcolor{lime}

\usepackage{amsmath}
\DeclareMathOperator*{\argmax}{argmax} 

\begin{document}
\title{VesselMorph: Domain-Generalized Retinal Vessel Segmentation via Shape-Aware Representation}
\titlerunning{VesselMorph}
\author{Dewei Hu\inst{1}\and
Hao Li\inst{1} \and
Han Liu\inst{2} \and
Xing Yao\inst{2} \and
Jiacheng Wang\inst{2} \and
Ipek Oguz\inst{1}\inst{2}}

\institute{Department of Electrical and Computer Engineering, Vanderbilt University\and 
Department of Computer Science, Vanderbilt University\\
\email{dewei.hu@vanderbilt.edu}}
\authorrunning{D. Hu et al.}

\maketitle              
\begin{abstract}
Due to the absence of a single standardized imaging protocol, domain shift between data acquired from different sites is an inherent property of medical images and has become a major obstacle for large-scale deployment of learning-based algorithms. For retinal vessel images, domain shift usually presents as the variation of intensity, contrast and resolution, while the basic tubular shape of vessels remains unaffected. Thus, taking advantage of such domain-invariant morphological features can greatly improve the generalizability of deep models. In this study, we propose a method named \textit{VesselMorph} which generalizes the 2D retinal vessel segmentation task by synthesizing a shape-aware representation. Inspired by the traditional Frangi filter and the diffusion tensor imaging literature, we introduce a Hessian-based bipolar tensor field to depict the morphology of the vessels so that the shape information is taken into account. We map the intensity image and the tensor field to a latent space for feature extraction. Then we fuse the two latent representations via a weight-balancing trick and feed the result to a segmentation network. We  evaluate on six public datasets of fundus and OCT angiography images from diverse patient populations. VesselMorph achieves superior generalization performance compared with competing methods in different domain shift scenarios.

\end{abstract}
\section{Introduction} \label{Sec:Intro}
\begin{figure}[t]
    \centering
    \begin{tabular}{ccccc}
        \includegraphics[width=.19\linewidth]{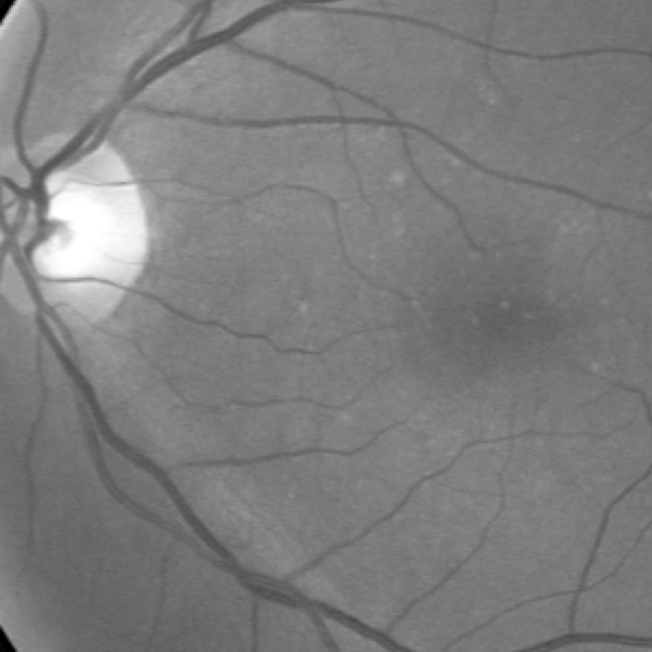}&
        \includegraphics[width=.19\linewidth]{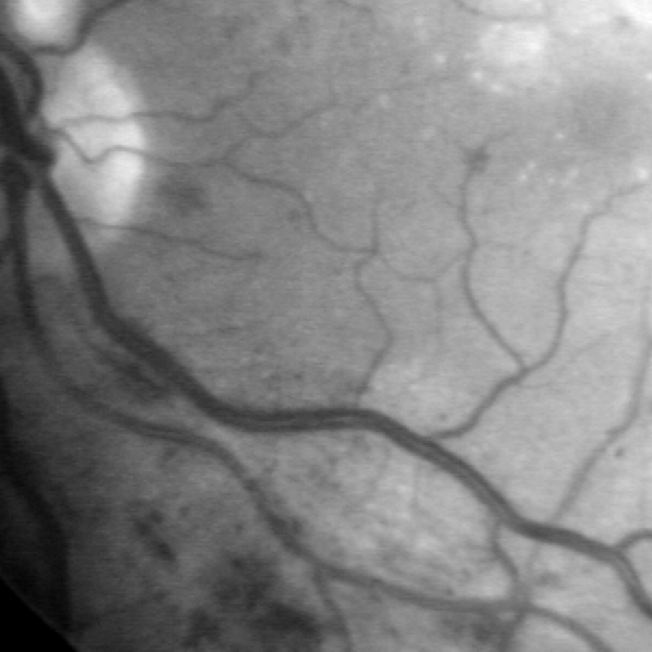}&
        \includegraphics[width=.19\linewidth]{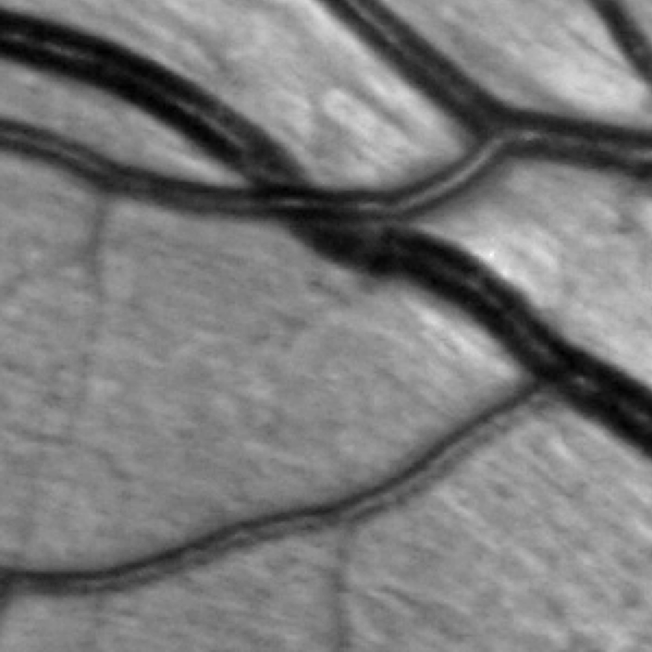}&
        \includegraphics[width=.19\linewidth]{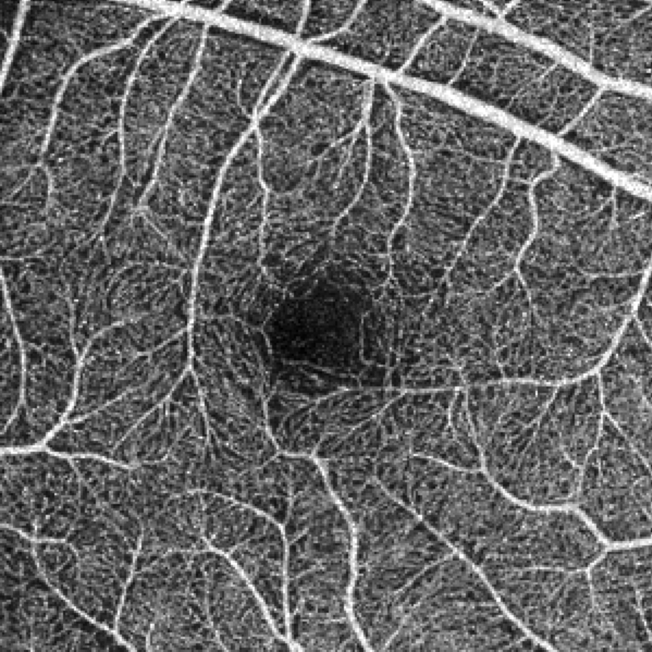}&
        \includegraphics[width=.19\linewidth]{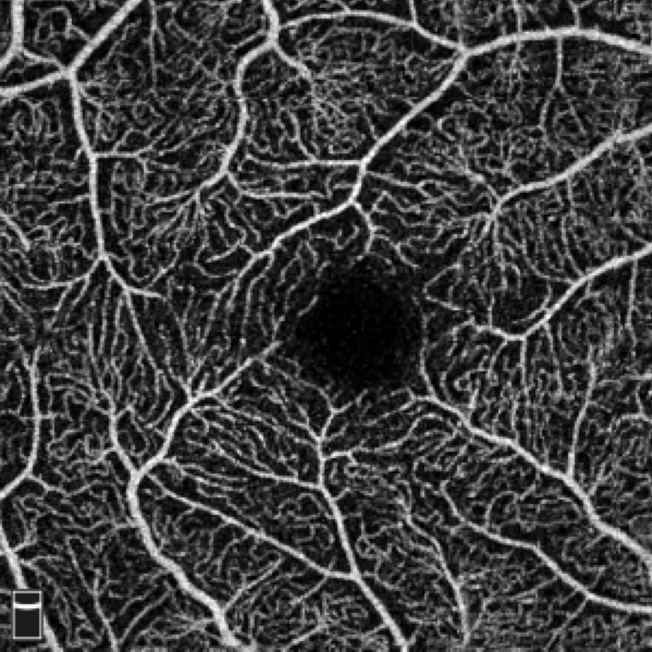}\\
    \end{tabular}
    \caption{Domain shift among retinal vessel images. Panels (1-3) are the green channel of fundus images. Panels (4-5) are OCT-A images. They all have the same size ($300\mathrm{pix}\times 300\mathrm{pix}$). Suppose (1) represents the source domain. Distribution shifts in test domain can be caused by (2) pathology, (3) resolution change, and (4, 5) different modality.}
    \label{fig:data_example}
\end{figure}

Medical images suffer from the distribution shift caused by the discrepancy in imaging acquisition protocols. Images can appear in different contrast, resolution and range of intensity values, even within the same modality. A set of examples is shown in Fig.\ \ref{fig:data_example}. This obstacle severely impedes the learning-based algorithms reaching clinical adoption. Therefore, much effort has been spent on solving the domain generalization (DG) problem so that the deep models can robustly work on out-of-distribution (OOD) data. There are three major types of solutions: data augmentation \cite{zhang2020generalizing,lyu2022aadg}, meta-learning \cite{finn2017model,li2018learning} and domain alignment \cite{zhou2022domain}. The first two strategies aim to improve the model's generalizability by either augmenting the source domain with additional data or replicating the exposure to OOD data during training. In contrast, the domain alignment strives to align the distribution of the target domains in either image \cite{he2021autoencoder} or feature space \cite{aslani2020scanner,li2020domain}.

We propose a novel method, \textit{VesselMorph}, to improve the DG performance by providing an explicit description of the domain-agnostic shape features as auxiliary training material. Even though traditional algorithms are outperformed by their learning-based counterparts in many aspects, they can typically  better generalize to any dataset, regardless of distribution shifts. Specifically for vessel segmentation, Frangi et al. \cite{frangi1998multiscale} proposed a Hessian-based model to express the tubular shape of vessels which can be regarded as a domain-invariant feature. Merging the Hessian-based shape description \cite{hu2021domain} with the principles of diffusion tensor imaging (DTI) \cite{le2001diffusion}, we introduce a bipolar tensor field (BTF) to explicitly represent the vessel shape by a tensor at each pixel. To effectively merge the features in the intensity image and the shape descriptor BTF, we employ a full-resolution feature extraction network to obtain an interpretable representation in the latent space from both inputs. This technique is broadly used in unsupervised segmentation \cite{liu2020variational,hu2021life} and representation disentanglement \cite{dewey2020disentangled,ouyang2021representation}. 

\begin{figure}[t]
    \centering
    \includegraphics[width=.81\linewidth]{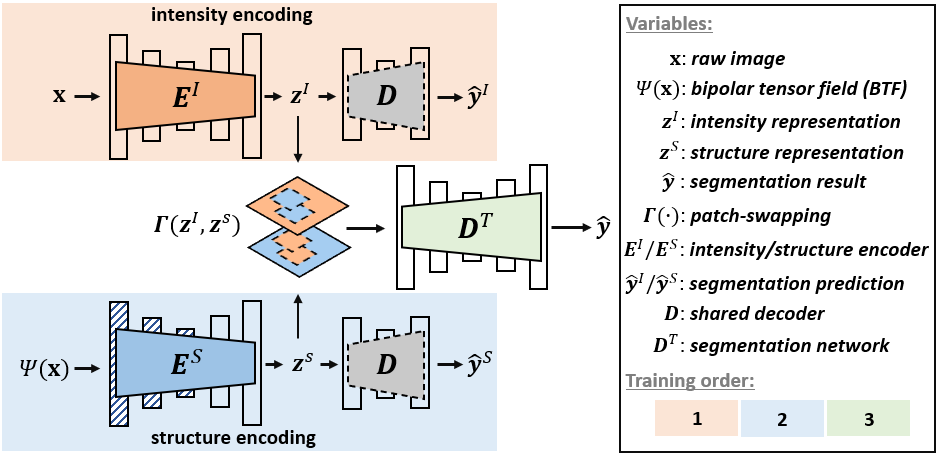}
    \caption{The overall structure of VesselMorph. The shaded layers include transformer blocks. The dashed line indicates $D$ will be discarded in testing.}
    \label{fig:pipeline}
\end{figure}

As shown in Fig.~\ref{fig:pipeline}, let $\mathbf{x}$ be the input image and $\Psi(\mathbf{x})$ the corresponding BTF. $D(E^I(\cdot))$ and $D(E^S(\cdot))$ are two feature extraction networks with a shared decoder $D$. We empirically observe that the intensity representation $\mathbf{z}^I$ can precisely delineate thinner vessels while the structure representation $\mathbf{z}^S$ works better on thick ones. We combine the strengths of the two pathways for a robust DG performance. The two latent images are fused by a weight-balancing trick $\Gamma(\mathbf{z}^I,\mathbf{z}^S)$ to avoid any potential bias induced by the selection of source domains. Finally, we train a segmentation network $D^T$ on the fused latent images. We compare the performance of VesselMorph to other DG approaches on four public datasets that represent various distribution shift conditions, and show that VesselMorph has superior performance in most OOD domains. Our contributions are:

\begin{itemize}
    \item[\ding{118}] A Hessian-based bipolar tensor field (BTF) that provides an explicit description of the vessel morphology (Sec.~\ref{Sec:BTF}).
    \item[\ding{118}] A full-resolution feature extraction network that generates vessel representation from both  the intensity image and the BTF (Sec.~\ref{Sec:vesselrepresentation}).
    \item[\ding{118}] A training pipeline that generates stable latent images for both pathways and a weight-balancing method to fuse the two representations (Sec.~\ref{Sec:fusion}).
    \item[\ding{118}] A comprehensive evaluation on public datasets which shows superior cross-resolution and cross-modality generalization performance (Sec.~\ref{Sec:Evaluation}).
\end{itemize}      

\section{Methods}

\subsection{Bipolar Tensor Field} \label{Sec:BTF}
Unlike ML models, our visual interpretation of vessels is rarely affected by data distribution shifts. Mimicking the human vessel recognition can thus help address the DG problem. In addition to intensity values, human perception of vessels also depends on the local contrast and the correlation in a neighborhood, which is often well described by the local Hessian. Inspired by the use of DTI to depict the white matter tracts, we create a Hessian-based bipolar tensor field to represent the morphology of vessels. Given a 2D input image $\mathbf{x}\in\mathbb{R}^{h\times w}$ and scale $\sigma$, the classical Frangi vesselness $\mathcal{V}(\sigma)$ \cite{frangi1998multiscale} is defined as:
\begin{equation}
\mathcal{V(\sigma)} = \begin{cases}
                      0 & \text{if }\lambda_2>0,\\
                      \exp\left(-\frac{\mathcal{R}_B^2}{2\beta^2}\right)\left[1-\exp\left(-\frac{S^2}{2c^2}\right)\right] & \text{else}
\end{cases}.
\end{equation}
Here, $\lambda_1, \lambda_2$ are the sorted eigenvalues of the Hessian $\mathcal{H}$,  $\mathcal{R}_B=\lambda_1/\lambda_2$, $S$ is the Frobenius norm of the Hessian ($\|\mathcal{H}\|_F$), $\beta=0.5$ and $c=0.5$. Note that we assume vessels are brighter than the background; fundus images are negated to comply. To represent vessels of different sizes, we leverage the multiscale vesselness filter that uses the optimal scale $\sigma^*$ for the Hessian  $\mathcal{H}(\mathbf{x}_{ij},\sigma)$ at each pixel $(i,j)$. This is achieved by grid search in the range $[\sigma_{min},\sigma_{max}]$ to maximize the vesselness $\mathcal{V}(\sigma)$, i.e., 
$\sigma^* = \argmax_{\sigma_{min}\leq \sigma\leq \sigma_{max}} \mathcal{V}(\sigma)$.
Then the optimized Hessian is represented by a $2\times 2$ matrix:
\begin{equation}
    \mathcal{H}(\mathbf{x}_{ij},\sigma^*)=(\sigma^*)^2\mathbf{x}_{ij} \ast \nabla^2 G(\mathbf{x}_{ij},\sigma^*)
\end{equation}
where $G(\mathbf{x}_{ij},\sigma^*)$ is a 2D Gaussian kernel with standard deviation $\sigma^*$. Then we apply the eigen decomposition to obtain the eigenvalues $\lambda_1$, $\lambda_2$ ($|\lambda_1|\leq|\lambda_2|$) and the corresponding eigenvectors $\mathbf{v}_1$, $\mathbf{v}_2$ at the optimal $\sigma^*$.

\begin{figure}[t]
\centering
\begin{tabular}{ccc}
\includegraphics[width=.26\linewidth]{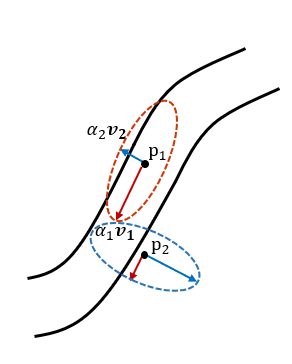}&
\includegraphics[width=.28\linewidth]{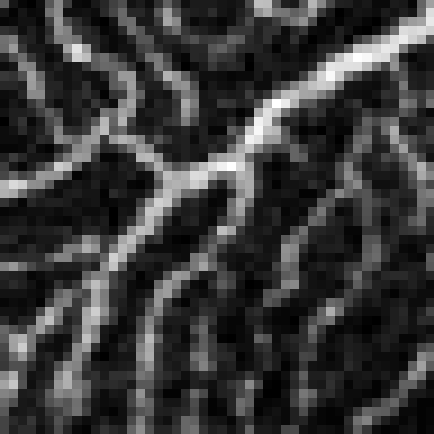}&
\includegraphics[width=.28\linewidth]{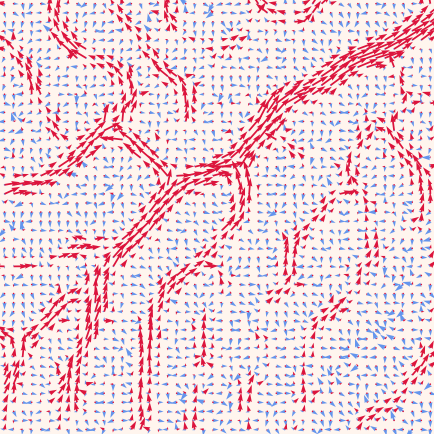}\\
\end{tabular}
\caption{\textbf{Left:} A simplified illustration of BTF. The red arrows indicate the orientation of $\mathbf{v}_1$ while the blue arrows correspond to $\mathbf{v}_2$. The ellipses represent the tensors at $p_1$ (in the vessel) and $p_2$ (in the background). \textbf{Right:} BTF applied on an OCTA image.}
\label{fig:BTF}
\end{figure}

Instead of solely analyzing the signs and magnitudes of the Hessian eigenvalues as in the traditional Frangi filter, we propose to leverage the eigenvectors along with custom-designed magnitudes to create our tensor field as shown in Fig.\ \ref{fig:BTF}(Left). The core idea of Frangi filter is to enhance the tubular structure by matching the vessel diameter with the distance between the two zero crossings in the second order derivative of Gaussian ( $2\sqrt{2}\sigma^*$). However, the solution is not guaranteed to land in range $[\sigma_{min},\sigma_{max}]$, especially for small vessels. Consequently, we observe that the inaccurate estimation of $\sigma^*$ results in a blurring effect at the vessel boundary, which is problematic for segmentation. As an example in Fig.~\ref{fig:BTF}(Left), the direction of $\mathbf{v}_1$ at $p_2$ aligns with that at $p_1$, even though $p_1$ is inside the vessel while $p_2$ is in the background but close to the boundary. This makes it difficult for the vector orientations alone to differentiate points inside and outside the vessel. To tackle this, we introduce the idea of a bipolar tensor by assigning a large magnitude to the orthogonal eigenvector $\mathbf{v}_2$ to points in the background, as shown in the blue dashed ellipse. Specifically, we define the magnitudes $\alpha_1$ and $\alpha_2$ associated with the eigenvectors $\mathbf{v}_1$ and $\mathbf{v}_2$ as:
\begin{align}
    \alpha_1 = 
    {\underbrace{%
     \vphantom{ \left(\frac{\lambda_1^2}{\|\mathcal{H}\|^2_{F}}\right) } 
     P(\mathbf{x}\leq \mathbf{x}_{ij})}_{\text{bright}}}
    {\underbrace{%
     \exp\left(-\epsilon\frac{\lambda_1^2}{\|\mathcal{H}\|^2_{F}}\right) 
    }_{\text{vessel-like}}} 
    ,\quad 
    \alpha_2 = 
    {\underbrace{%
     \vphantom{ \left(\frac{\lambda_2^2}{\|\mathcal{H}\|^2_{F}}\right) } 
     P(\mathbf{x}> \mathbf{x}_{ij})}_{\text{dark}}}
    {\underbrace{%
     \exp\left(-\epsilon\frac{\lambda_2^2}{\|\mathcal{H}\|^2_{F}}\right) 
    }_{\text{vessel-like}}}
\end{align}
where $P(\mathbf{x} > \mathbf{x}_{ij})$ is the probability that the intensity of a random pixel $x$ in the image is greater than $\mathbf{x}_{ij}$. This is equivalent to normalizing the histogram by the factor $hw$ and computing the cumulative distribution function at $\mathbf{x}_{ij}$. This term thus provides a normalized brightness function in the range $[0,1]$. The exponential term represents how vessel-like the voxel is by using a normalized eigenvalue, and is in the $[0,1]$ range as well. $\epsilon$ is a constant that controls the sensitivity, which is empirically set to $0.5$. With the custom magnitudes $\alpha_1$ and $\alpha_2$, the two poles can better differentiate vessels from the background. Fig.~\ref{fig:BTF}(Right) is an example of BTF on an OCTA image. In practice, we stack the two vectors as the input to the structural encoding network, i.e., $\Psi(\mathbf{x}_{ij})=\left[\alpha_1\mathbf{v}_1^\top,\alpha_2\mathbf{v}_2^\top\right]^\top \in \mathbb{R}^{4\times 1}$.

\subsection{Latent Vessel Representation} \label{Sec:vesselrepresentation}
Preserving the spatial resolution for the bottleneck of models with U-Net backbone is a common strategy to emphasize the structural features in unsupervised segmentation \cite{liu2020variational,hu2021life} and representation disentanglement \cite{dewey2020disentangled,ouyang2021representation}. We employ a network that has a full-resolution ($h\times w$ pixels) latent space as the feature extraction model. We propose to extract vessel structure from both the intensity image $\mathbf{x}\in \mathbb{R}^{h\times w}$ and its corresponding BTF, $\Psi(\mathbf{x})\in \mathbb{R}^{4\times h\times w}$. Therefore, in Fig.~\ref{fig:pipeline}, the intensity $D(E^I(\cdot))$ and structure $D(E^S(\cdot))$ encoding pathways share the decoder D, and the latent images $\mathbf{z}^I,\mathbf{z}^S \in \mathbb{R}^{h\times w}$. To distribute more workload on the encoder, $D$ has a shallower architecture and will be discarded in testing. For the intensity encoding, the model is optimized by minimizing the segmentation loss function defined as the combination of cross-entropy and Dice loss:
\begin{equation}
\label{loss:segloss}
\mathcal{L}_{seg}=-\frac{1}{N}\sum_{n=1}^N\mathbf{y}_n\log\hat{\mathbf{y}}^I_n + \left(1-\frac{2\sum_{n=1}^N \mathbf{y}_n\Hat{\mathbf{y}}^I_n}{\sum_{n=1}^N \mathbf{y}_{n}^{2}+(\hat{\mathbf{y}}^I_n)^2}\right)
\end{equation}
where $N=h\times w$ is the total number of pixels in the image, $\mathbf{y}$ is the ground truth and $\hat{\mathbf{y}}^I$ is the prediction from the training-only decoder $D$. Although there is no explicit constraint on the latent image $E^I(\mathbf{x})=\mathbf{z}^I$, we note that the segmentation-based supervision encourages it to include the vessels while most other irrelevant features are filtered out. Hence, we can view the latent feature as a vessel representation. 

Our approach is slightly different for the structure encoding as we observe that it is hard for the feature extraction network to generate a stable latent image that is free of artifacts when the number of input channels is greater than 1. Thus, it is necessary to use $E^I$ as a teacher model that provides direct supervision on the vessel representation. In other words, we first train the intensity encoding path to get $E^{I}$ and $D$, then train the $E^{S}$ by leveraging both the segmentation loss in Eq.~\ref{loss:segloss} and a similarity loss defined as: 
\begin{equation}
    \mathcal{L}_{sim}(\mathbf{z}^S,\mathbf{z}^I) =   \sum_{n=1}^N\|\mathbf{z}^S_{n}-\mathbf{z}^I_{n}\|_1 + \mathrm{SSIM}(\mathbf{z}^S, \mathbf{z}^I)
\label{loss:simloss}
\end{equation}
which is a weighted sum of $L_1$ norm and structural similarity loss SSIM \cite{hore2010image}. 
SSIM is defined as 
$
    \mathrm{SSIM}(A, B) = 2\frac{(2\mu_{A}\mu_{B}+c_1)(2\sigma_{AB}+c_2)}{(\mu_{A}^2+\mu_{B}^2+c_1)(\sigma_{A}^2+\sigma_{B}^2+c_2)},
$
where $\mu$ and $\sigma$ represent the mean and standard deviation of the image, and we set $c_1=0.01$ and $c_2=0.03$. The overall loss function for the structural encoding is thus
$\mathcal{L}(\Psi(\mathbf{x}),\mathbf{y}) = \omega_1 \mathcal{L}_{seg}(\hat{\mathbf{y}}^S,\mathbf{y})+\omega_2 \mathcal{L}_{sim}(\mathbf{z}^S,\mathbf{z}^I)$,
with empirically determined weights $\omega_1=1$, $\omega_2=5$. Experimentally, we found that the $\mathbf{z}^I$ is good at preserving small vessels, while $\mathbf{z}^S$ works better on larger ones.

\subsection{Fusion of Vessel Representations} \label{Sec:fusion}
Given the two synthesized vessel representations $\mathbf{z}^I$ and $\mathbf{z}^S$, we need to introduce a fusion method to take advantage of both intensity and structure features. Naively stacking these two channels as input to the segmentation network is prone to inducing bias: if $\mathbf{z}^I$ is consistently better for images from the source domain, then the downstream task model $D^{T}$ would learn to downplay the contribution of $\mathbf{z}^S$ due to this biased training data. As a result, despite its potential to improve performance, $\mathbf{z}^S$ would be hindered from making a significant contribution to the target domain during testing. To circumvent this issue, we propose a simple weight-balancing trick. As illustrated in Fig.~\ref{fig:pipeline}, we randomly swap some patches between the two latent images so that $D^{T}$ does not exclusively consider the feature from a single channel, even for biased training data. This trick is feasible because $\mathbf{z}^S$ and $\mathbf{z}^I$ are in the same intensity range, due to the similarity constraints applied in Eq.~\ref{loss:simloss}. Thus the input to $D^{T}$ is $\Tilde{\mathbf{x}} = \Gamma(\mathbf{z}^I,\mathbf{z}^S)$, where $\Tilde{\mathbf{x}}\in \mathbb{R}^{2\times h\times w}$. The loss function leveraged for $D^{T}$ is the same as Eq.~\ref{loss:segloss}.

\SetKwInOut{Input}{input}
 \SetKwInOut{Output}{output}
\begin{algorithm}[H]
\label{alg1}
  \DontPrintSemicolon
  \Input{Source domains $\mathcal{S}=\{(\mathbf{x}_i,\mathbf{y}_i)\}_{i=1}^K$,
  \\ hyperparameters: $\epsilon, \sigma_{min}, \sigma_{max}$, $\eta_{E^I}$, $\eta_{E^S}$, $\eta_{D^T}$ $\omega_1$, $\omega_2$}
  \Output{parameters of models $\theta_I^*$, $\theta_S^*$, $\varphi^*_T$} 
  
   \textcolor{gray}{\tcp{Train the intensity encoder $E^I$ as a teacher model}}
   \nl\Repeat{converge}{ 
    \For{$i=1:K$}{
        $\theta_I^{'} \leftarrow \theta_I - \eta_{E^I}(i) \nabla\mathcal{L}_{seg}(D(E^{I}(\mathbf{x}_i)),\mathbf{y}_i)$ \; 
        }
  } 
  \nl Generate the tensor field: $\Psi(\mathbf{x})$ \;
  \textcolor{gray}{\tcp{Train the structure encoder $E^S$ as a student model}}
  \nl\Repeat{converge}{
    \For{$i=1:K$}{
        $\hat{\mathbf{y}}_i \leftarrow D(E^{S}(\Psi(\mathbf{x}_i)))$ \;
        $\mathcal{L}(\Psi(\mathbf{x}_i),\mathbf{y}_i) \leftarrow \omega_1 \mathcal{L}_{seg}(\hat{\mathbf{y}}_i,\mathbf{y}_i)+\omega_2 \mathcal{L}_{sim}(E^{S}(\Psi(\mathbf{x}_i)),E^{I}(\mathbf{x}_i))$ \;
        $\theta_S^{'} \leftarrow \theta_S - \eta_{E^S}(i) \nabla\mathcal{L}(\Psi(\mathbf{x}_i),\mathbf{y}_i)$\; 
        }
  } 
  
  \textcolor{gray}{\tcp{Train the segmentation network $D^{T}$}}
  \nl\Repeat{converge}{
    \For{$i=1:K$}{
        $\tilde{\mathbf{x}}_i \leftarrow \Gamma(E^{I}(\mathbf{x}_i),E^{S}(\Phi(\mathbf{x}_i)))$ \; 
        $\varphi^{'}_T \leftarrow \varphi_T - \eta_{D^T}(i) \nabla\mathcal{L}_{seg}(D^{T}(\tilde{\mathbf{x}}_i),\mathbf{y}_i)$ \; 
        }
    }
  \caption{Training of VesselMorph}
\end{algorithm}

The complete algorithm for training of VesselMorph is shown in Algorithm.\ref{alg1}. Briefly, we first train the intensity encoder $E^I$ as it is easier to generate a stable vessel representation $\mathbf{z}^I$. Then a structure encoder $E^S$ is trained with the supervision of the ground truth and teacher model $E^I$ so that an auxiliary representation $\mathbf{z}^S$ is extracted from the structural descriptor BTF. The last step is to train a segmentation network $D^{T}$ with the fusion of the two vessel maps $\Gamma(\mathbf{z}^I,\mathbf{z}^S)$. During testing, the patch-swapping is no longer needed, so we simply concatenate $E^I(\mathbf{x})$ and $E^S(\Psi(\mathbf{x}))$ as the input to $D^T$.

\begin{table}[t]
\centering
\scalebox{0.93}{
\begin{tabular}{p{0.3\textwidth}>{\centering}
                p{0.16\textwidth}>{\centering}
                p{0.16\textwidth}>{\centering\arraybackslash}
                p{0.16\textwidth}>{\centering\arraybackslash}
                p{0.12\textwidth}}
\specialrule{.1em}{.05em}{.05em}
 & \small{\textbf{modality}} & \small{\textbf{resolution}} & \footnotesize{$\#$} \small{\textbf{sample}} & \small{\textbf{domain}} \\
\hline
\footnotesize{DRIVE \cite{staal2004ridge}} & \small{fundus} & \small{$565\times 584$} & \small{20} & \small{$\mathcal{S}$}\\
\footnotesize{STARE \cite{hoover2000locating}} & \small{fundus} & \small{$700\times 605$} & \small{20} & \small{$\mathcal{S}$}\\
\footnotesize{ARIA \cite{farnell2008enhancement}} & \small{fundus} & \small{$768\times 576$} & \small{$61/59/23$} & \small{$\mathcal{S}/\mathcal{T}/\mathcal{T}$}\\
\rowcolor{Gray1}
\footnotesize{{HRF} \cite{budai2013robust}} & \small{fundus} & \small{$3504\times 2336$} & \small{$15/15/15$} & \small{$\mathcal{T}$}\\
\rowcolor{Gray2}
\footnotesize{{OCTA-500(6M)} \cite{li2020image}} & \small{OCTA} & \small{$400\times 400$} & \small{300} & \small{$\mathcal{T}$}\\
\rowcolor{Gray2}
\footnotesize{{ROSE} \cite{ma2020rose}} & \small{OCTA} & \small{$304\times 304$} & \small{30} & \small{$\mathcal{T}$}\\
\specialrule{.1em}{.05em}{.05em}
\end{tabular}}
\caption{Publicly available datasets used in our experiments. For ARIA and HRF, we list the number of samples per class. ARIA classes: healthy, diabetic and AMD (age-related macular degeneration). HRF classes: healthy, diabetic and glaucoma. The shading of the rows indicates datasets in similar distributions to each other.}
\label{tab:dataset}
\end{table}

\section{Experiments}
\subsection{Experimental Settings} 

\paragraph{\normalfont{\textbf{\underline{Datasets.}}}}The 6 publicly available datasets used in this study are listed in Table \ref{tab:dataset}. Since there are more labeled fundus data available, we set up a source domain $\mathcal{S}$ that includes three fundus datasets: DRIVE, STARE and the control subjects in ARIA. In the target domain $\mathcal{T}$, we test the performance of the model under three different conditions: pathology (diabetic/AMD subjects in ARIA), resolution change (HRF) and cross-modality (OCTA500 and ROSE).

\noindent{\normalfont{\textbf{\underline{Compared methods.}}}} We pick one representative algorithm from each of the three major categories of DG approaches (Sec.~\ref{Sec:Intro}) as a competing method. For data augmentation, we implement BigAug \cite{zhang2020generalizing}. For meta-learning, we use the MASF \cite{dou2019domain} model. For domain alignment, we use the domain regularization network \cite{aslani2020scanner}. In addition, we also include VFT \cite{hu2021domain} which proposes the idea of shape description for DG. The baseline model is a vanilla residual U-Net trained on $\mathcal{S}$, and the oracle model is the same network trained directly on each target domain to represent the optimal performance. Note that for a fair comparison, we set the baseline model to have a bit more parameters than $D(E^I(\cdot))$ ($7.4\times 10^5:6.7\times 10^5$). 

\input{qualFig.tex}

\noindent{\normalfont{\textbf{\underline{Implementation Details.}}}} We use the residual U-Net structure for $E^I$, $D$ and $D^T$. To take advantage of the tensor field, the structure encoder $E^S$ is equipped with parallel transformer blocks with different window sizes as proposed in \cite{hu2021domain}. All networks are trained and tested on an NVIDIA RTX 2080TI 11GB GPU. We use a batch size of 5 and train for 100 epochs. We use the Adam optimizer with the initial learning rate $\eta_{E^I} = \eta_{E^S} = 5\times 10^{-4}$, $\eta_{D^T} = 1\times 10^{-3}$, decayed by 0.5 for every 3 epochs. For fundus images, we use the green channel as network input $\mathbf{x}$. The intensity values are normalized to $[0,1]$.

\subsection{Results} \label{Sec:Evaluation}

\begin{figure}[t]
    \centering
    \begin{tabular}{ccc}
         \includegraphics[width=.32\linewidth]{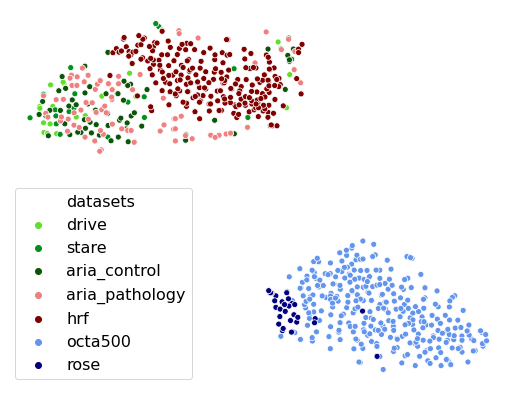}&
         \includegraphics[width=.32\linewidth]{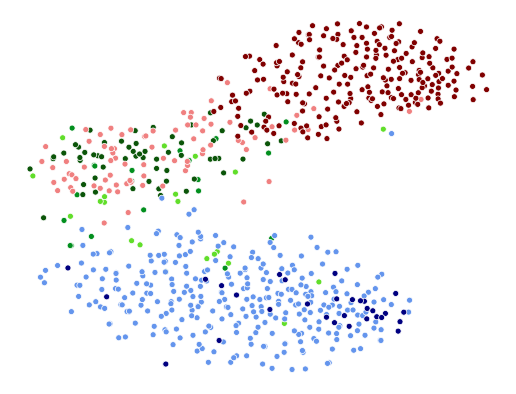}&
         \includegraphics[width=.32\linewidth]{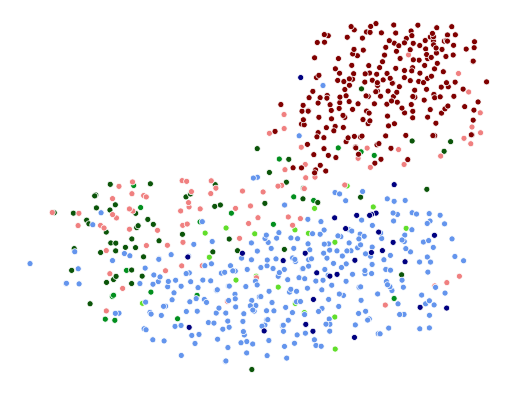}\\
    \end{tabular}
    \caption{t-SNE on raw data $\mathbf{x}$(left), $\mathbf{z}^I$(center) and $\mathbf{z}^S$(right). $\mathcal{S}$ is coded by shades of green, while fundus and OCTA in $\mathcal{T}$ are coded by red and blue shades respectively. Both intensity and structure representations reduce the domain gaps between datasets.}
    \label{fig:TSNE}
\end{figure}
\input{table2.tex}

\begin{figure}[t]
    \centering
    \includegraphics[width=.98\linewidth]{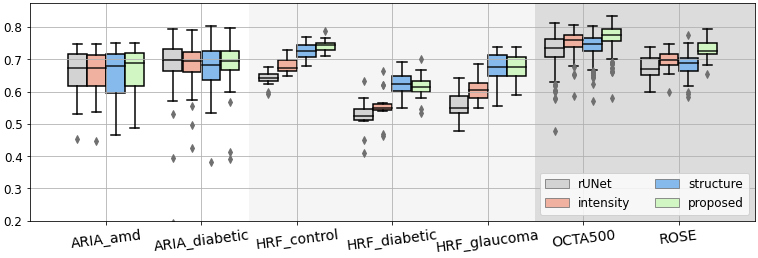}
    \caption{Quantitative ablation results. Dice scores of vanilla residual U-Net, intensity encoding $D(\mathbf{z}^I)$, structural encoding $D(\mathbf{z}^S)$ and the final output $D^T(\Gamma(\mathbf{z}^I,\mathbf{z}^S))$.The background is encoded the same way as Table 1. We note that the $\mathbf{z}^S$ is especially useful in capturing the thick vessels in HRF, whereas $\mathbf{z}^I$ provides additional precision in the thin vessels in the OCTA datasets. The proposed model combines these advantages and is robust across the board.}
    \label{fig:ablation_boxplot}
\end{figure}

Fig.\ \ref{tab:segment} shows a qualitative ablation study: it illustrates that the intensity representation $\mathbf{z}^I$ may miss large vessels in the very high-resolution HRF images, while $\mathbf{z}^S$ remains robust. In contrast, $\mathbf{z}^I$ provides sharper delineation for very thin vessels in ROSE. The fusion of both pathways outperforms either pathway for most scenarios. These observations are further supported by the quantitative ablation study in Fig.\ref{fig:ablation_boxplot}. We note that $\mathbf{z}^S$ and $\mathbf{z}^I$ can be used as synthetic angiograms that provide both enhanced vessel visualization and model interpretability. 

Fig.\ \ref{fig:TSNE} shows the t-SNE plots \cite{van2008visualizing} of the datasets. The distribution gaps between datasets are greatly reduced for the two latent vessel representations.

Table \ref{tab:result} compares all methods on the target domain $\mathcal{T}$. For the diseased ARIA data, all methods show comparable performance and are not significantly different from the baseline. VesselMorph has the best OOD outcome for both cross-modality (dark gray) and cross-resolution (light gray) scenarios, except the OCTA500 dataset where VFT, MASF and VesselMorph perform similarly. The results of VFT and VesselMorph prove the value of the shape information.

\section{Conclusion}
In this work, we propose to solve the DG problem by explicitly modeling the domain-agnostic tubular vessel shape with a bipolar tensor field which connects traditional algorithms with deep learning. We extract vessel representation from both intensity and BTF, then fuse the information from the two pathways so that the segmentation network can better exploit both types of description. Our VesselMorph model provides significant quantitative improvement on Dice score across a variety of domain shift conditions, and its latent images offer enhanced vessel visualization and interpretability.   

\noindent{\normalfont{\textbf{Acknowledgements.}}} This work is supported by the NIH grant R01EY033969 and the Vanderbilt University Discovery Grant Program.

\bibliographystyle{splncs04}
\bibliography{refs.bib}
\end{document}

%% file: qualFig.tex
\newcommand{\rsize}{0.160}
\setlength{\tabcolsep}{0.5pt}
\renewcommand{\arraystretch}{0.5}

\begin{figure}[t]
    \centering
    \begin{tabular}{ccc|ccc}
        \multicolumn{3}{c|}{HRF ${\scriptstyle[1000\times 1000]}$} & \multicolumn{3}{c}{ROSE ${\scriptstyle[200\times 200]}$} \\
        $\mathbf{x}$ & $\mathbf{z}^I$ & $\mathbf{z}^S$ & 
        $\mathbf{x}$ & $\mathbf{z}^I$ & $\mathbf{z}^S$ \\
        \specialrule{.1em}{.05em}{.05em}
        \includegraphics[width=\rsize\linewidth]{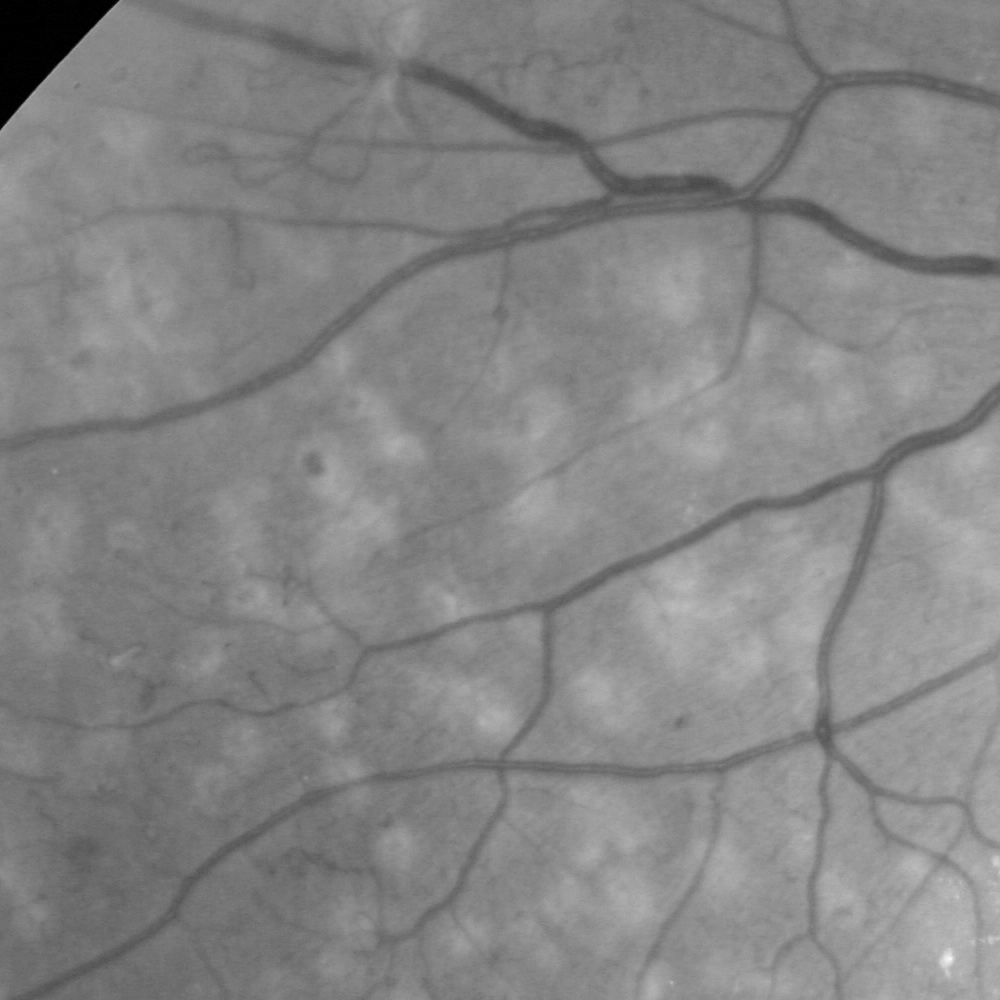} & \includegraphics[width=\rsize\linewidth]{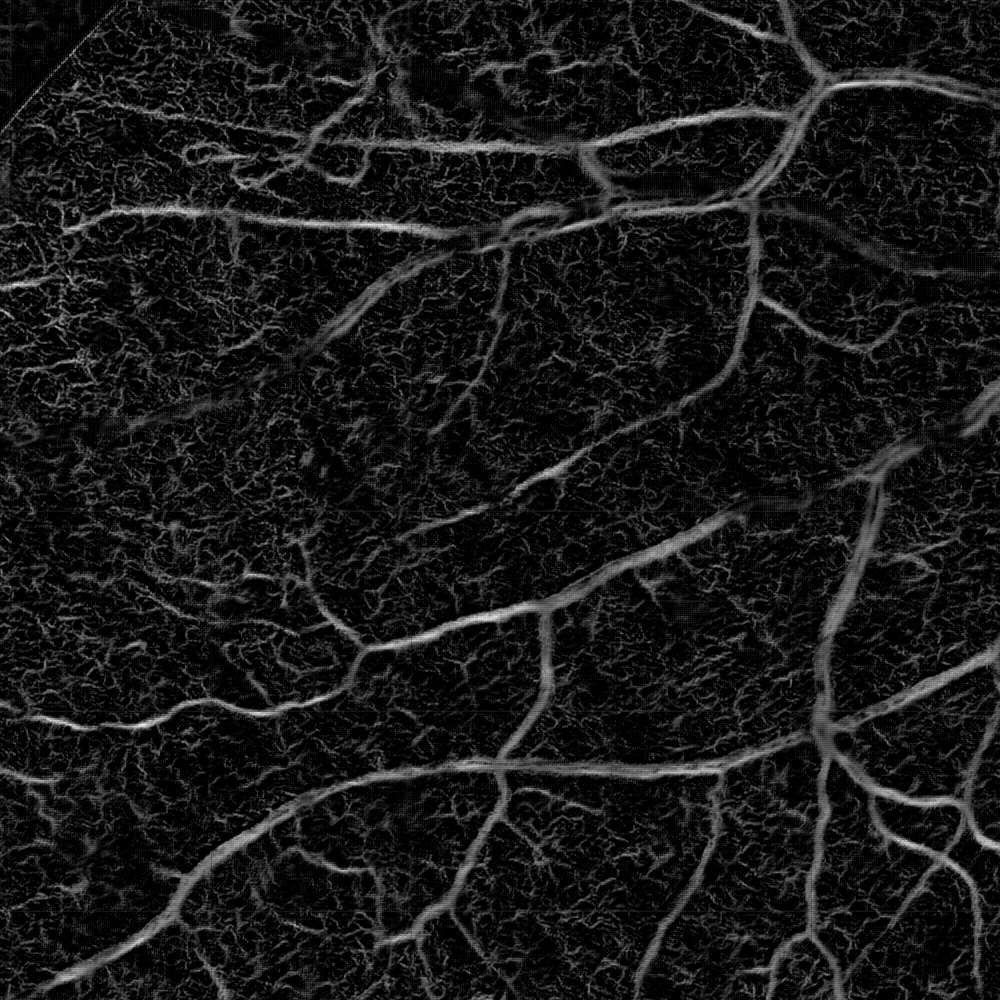} &
        \includegraphics[width=\rsize\linewidth]{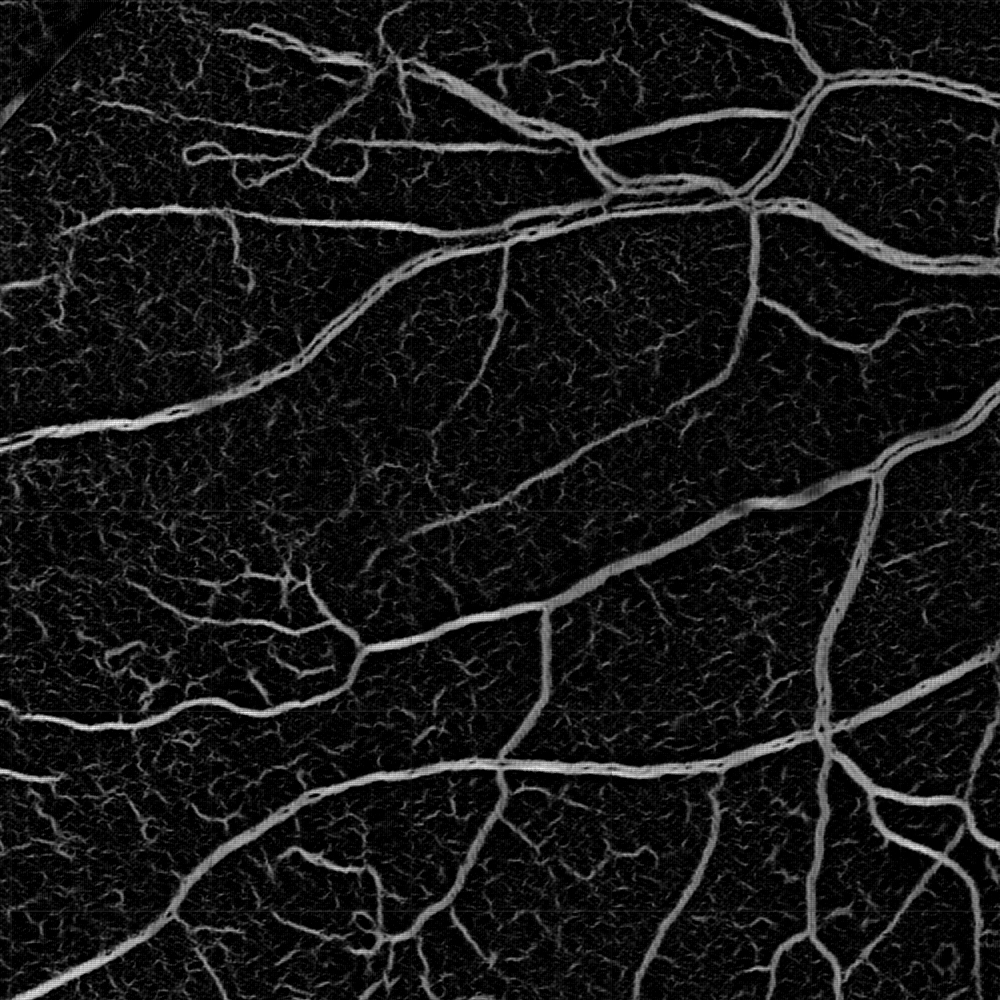} & \includegraphics[width=\rsize\linewidth]{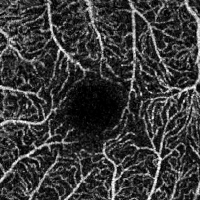} & \includegraphics[width=\rsize\linewidth]{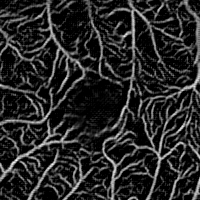} &
        \includegraphics[width=\rsize\linewidth]{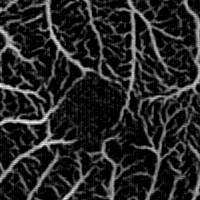} \\
        
        \includegraphics[width=\rsize\linewidth]{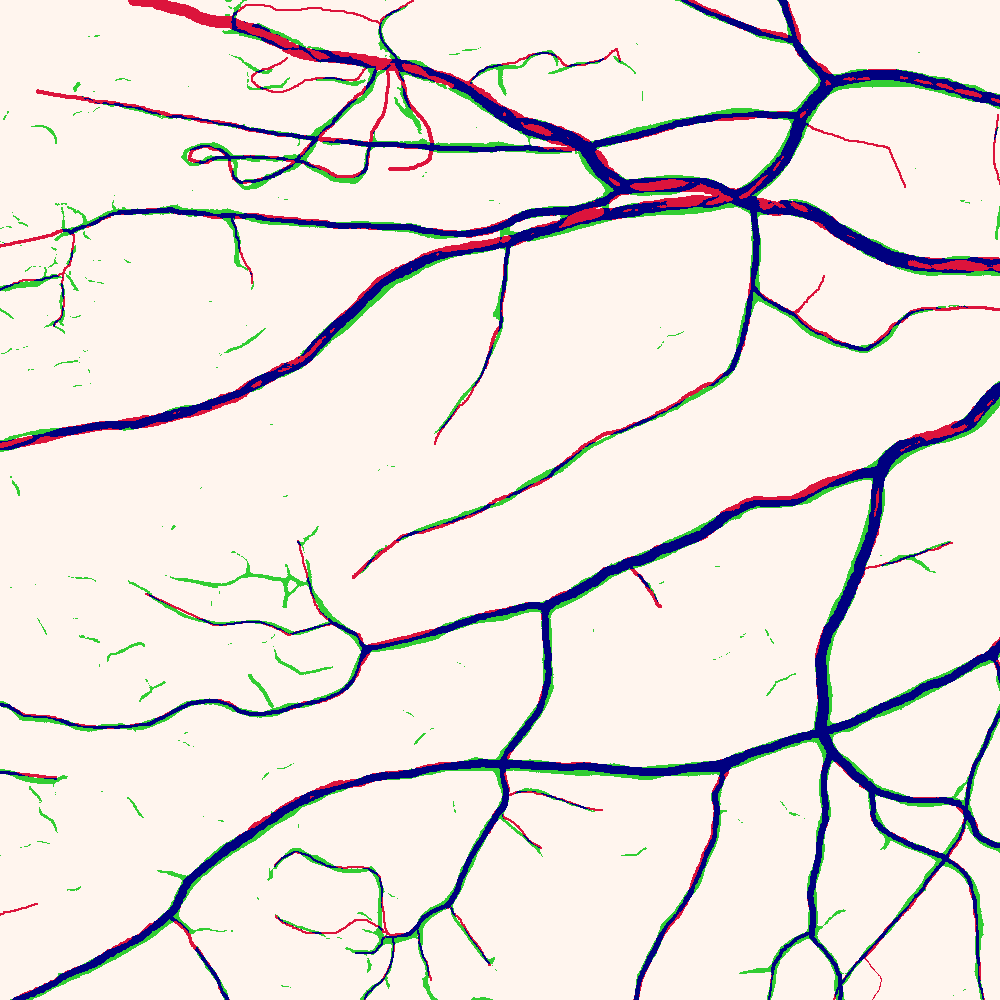} & \includegraphics[width=\rsize\linewidth]{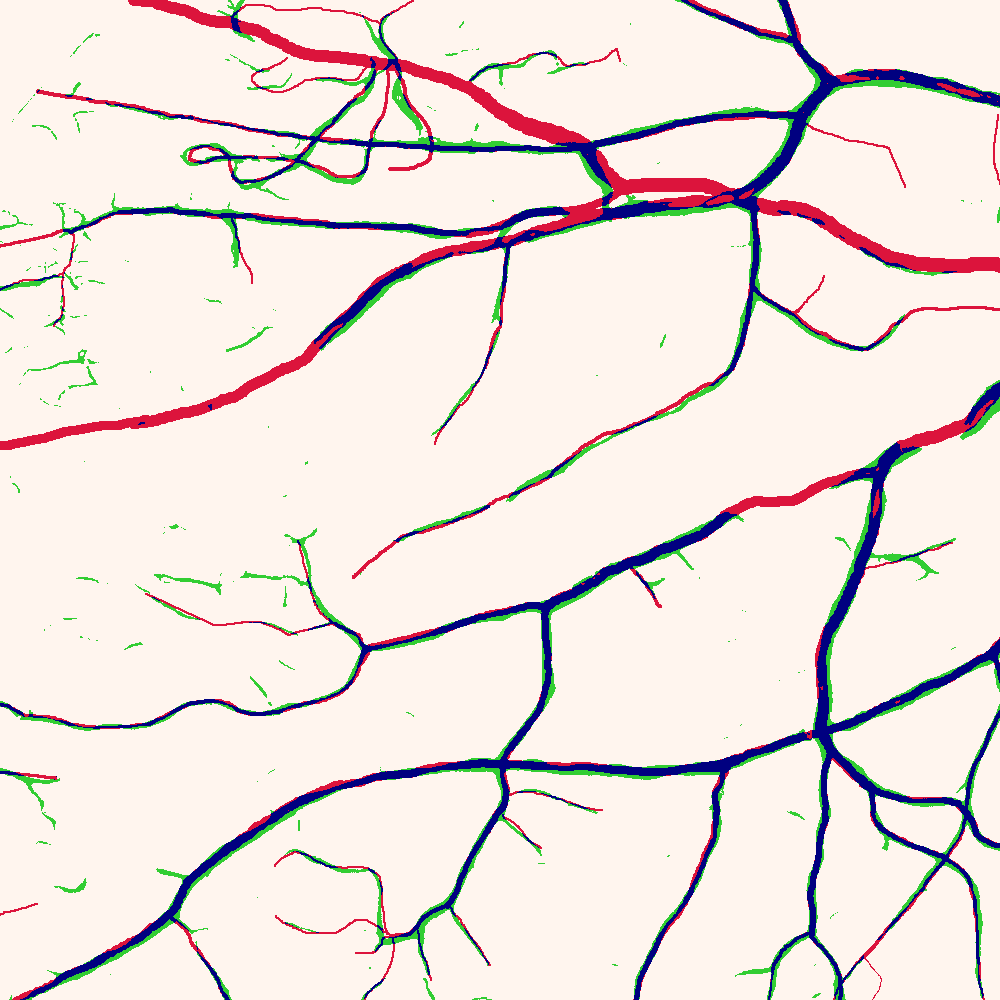} &
        \includegraphics[width=\rsize\linewidth]{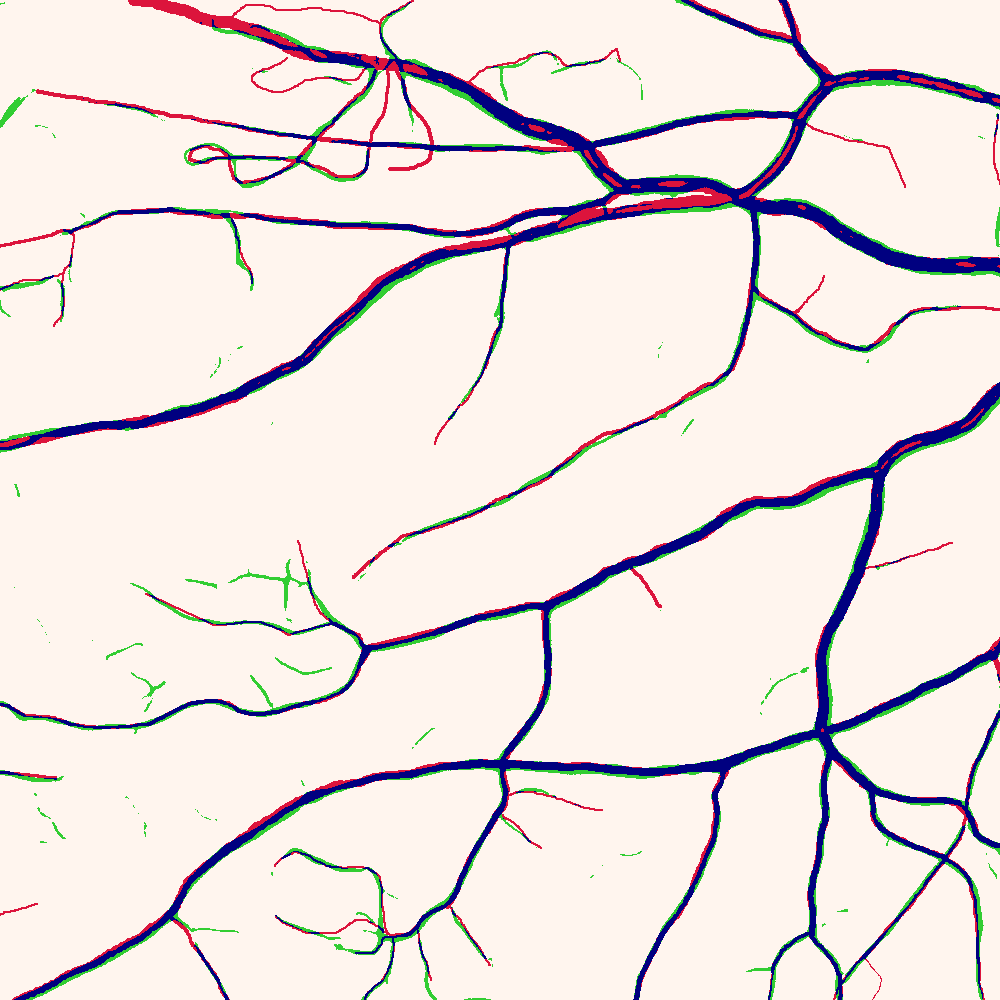} & \includegraphics[width=\rsize\linewidth]{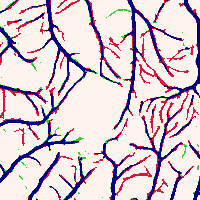} & \includegraphics[width=\rsize\linewidth]{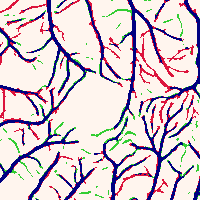} &
        \includegraphics[width=\rsize\linewidth]{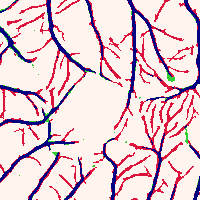} \\
    \end{tabular}
    \caption{Qualitative ablation. The shown patches are $1000\times 1000$pix for HRF diabetic image and $200\times 200$pix for ROSE. \textbf{Top row:} raw image, $\mathbf{z}^I$ and $\mathbf{z}^S$. \textbf{Bottom row:} the VesselMorph segmentation and prediction from each pathway, i.e., $D^T(\Gamma(\mathbf{z}^I,\mathbf{z}^S))$, $D(\mathbf{z}^I)$, and $D(\mathbf{z}^S)$. \textbf{Red} and \textbf{green} indicate the false negative (FN) and false positive (FP), respectively. $\mathbf{z}^I$ may miss large vessels, while $\mathbf{z}^S$ may miss thin ones. The fusion provides robust performance, as can also be seen quantitatively in Supp.\ Fig.\ 1.}
    \label{tab:segment}
\end{figure}

%% file: table2.tex
\renewcommand{\arraystretch}{0.95}
\newcolumntype{a}{>{\columncolor{Gray1}}p}
\newcolumntype{b}{>{\columncolor{Gray2}}p}
\begin{table}[t]
    \centering
    \begin{tabular}{p{0.13\textwidth}>{\centering}
                    p{0.12\textwidth}>{\centering\arraybackslash}
                    p{0.12\textwidth}>{\centering\arraybackslash}
                    a{0.12\textwidth}>{\centering\arraybackslash}
                    a{0.12\textwidth}>{\centering\arraybackslash}
                    a{0.12\textwidth}>{\centering\arraybackslash}
                    b{0.12\textwidth}>{\centering\arraybackslash}
                    b{0.12\textwidth}}
        \specialrule{.1em}{.05em}{.05em}
        \hspace{0.2em} \multirow{2}{*}{\small Method} &  \multicolumn{2}{c}{\scriptsize ARIA} & \multicolumn{3}{c}{\cellcolor{Gray1}{\scriptsize HRF}} & & \\
         & \scriptsize amd & \scriptsize diabetic & \scriptsize healthy & \scriptsize diabetic & \scriptsize glaucoma & \multirow{-2}{*}{\scriptsize OCTA 500} & \multirow{-2}{*}{\scriptsize ROSE}\\
        \specialrule{.1em}{.05em}{.05em}
        \scriptsize \hspace{0.4em} \textit{baseline} & $\scalemath{0.9}{0.6382}$ & $\scalemath{0.9}{0.6519}$ & $\scalemath{0.9}{0.6406}$ & $\scalemath{0.9}{0.5267}$ & $\scalemath{0.9}{0.5566}$ & $\scalemath{0.9}{0.7316}$ & $\scalemath{0.9}{0.6741}$ \\
        \hline
        \scriptsize \hspace{0.3em} \texttt{+}Regular & $\scalemath{0.9}{0.6489}$ & $\scalemath{0.9}{0.6697}$ & $\scalemath{0.9}{0.6403}$ & $\scalemath{0.9}{0.5216}$ & $\scalemath{0.9}{0.5625}$ & $\scalemath{0.9}{0.7354}$ & $\scalemath{0.9}{0.6836}$ \\
        \scriptsize \hspace{0.3em} \texttt{+}BigAug & $\scalemath{0.9}{\underline{0.6555}}$ & $\scalemath{0.9}{0.6727}$ & $\scalemath{0.9}{0.6613}$ & $\scalemath{0.9}{0.5389}$ & $\scalemath{0.9}{0.5735}$ & $\scalemath{0.9}{0.7688}$ & $\scalemath{0.9}{0.6932}$ \\
        \scriptsize \hspace{0.3em} \texttt{+}MASF & $\scalemath{0.9}{0.6533}$ & $\scalemath{0.9}{\underline{0.6775}}$ & $\scalemath{0.9}{0.6131}$ & $\scalemath{0.9}{0.5358}$ & $\scalemath{0.9}{0.5629}$ & $\scalemath{0.9}{\underline{0.7765}}$ & $\scalemath{0.9}{0.6725}$ \\
        \scriptsize \hspace{1.1em} VFT & $\scalemath{0.9}{0.6181}$ & $\scalemath{0.9}{0.6405}$ & $\scalemath{0.9}{\underline{0.7058}}$ & $\scalemath{0.9}{\underline{0.5732}}$ & $\scalemath{0.9}{\underline{0.6410}}$ & $\scalemath{0.9}{\bf{0.7791}}$ & $\scalemath{0.9}{\underline{0.7281}}$ \\
        \scriptsize VesselMorph & $\scalemath{0.9}{\bf{0.6619}}^{\sim}$ & $\scalemath{0.9}{\bf{0.6787}}^{\sim}$ & $\scalemath{0.9}{\bf{0.7420}}^{\dagger}$ & $\scalemath{0.9}{\bf{0.6145}}^{\dagger}$ & $\scalemath{0.9}{\bf{0.6756}}^{\dagger}$ & $\scalemath{0.9}{0.7714}^{\dagger}$ & $\scalemath{0.9}{\bf{0.7308}}^{\dagger}$ \\
        \hline
        \scriptsize \hspace{0.7em} \textit{oracle} & $\scalemath{0.9}{\textcolor{gray}{0.7334}}$ & $\scalemath{0.9}{\textcolor{gray}{0.7065}}$ & $\scalemath{0.9}{\textcolor{gray}{0.8358}}$ & $\scalemath{0.9}{\textcolor{gray}{0.7524}}$ & $\scalemath{0.9}{\textcolor{gray}{0.7732}}$ & $\scalemath{0.9}{\textcolor{gray}{0.8657}}$ & $\scalemath{0.9}{\textcolor{gray}{0.7603}}$\\
        \specialrule{.1em}{.05em}{.05em}
    \end{tabular}
    \caption{Dice values for testing on target domains. \textbf{Boldface}: best result. \underline{Underline}: second best result. $^\sim: \text{p-value} \geq 0.05$, $^\dagger: \text{p-value} \ll 0.05$ in paired t-test against the baseline output. The background is color-coded the same way as Table \ref{tab:dataset}.}
    \label{tab:result}
\end{table}